# Large scale digital prostate pathology image analysis combining feature extraction and deep neural network

Naiyun Zhou[1], Andrey Fedorov[2,3], Fiona Fennessy[2,3,4], Ron Kikinis[2,3,5,6], Yi Gao[1,*]

[1]Stony Brook University, Stony Brook, NY, U.S.A.
[2]Brigham and Women's Hospital, Boston, MA, U.S.A.
[3]Harvard Medical School, Boston, MA, U.S.A.
[4]Dana-Farber Cancer Institute, Boston, MA, U.S.A.
[5]Fraunhofer Institute for Medical Image Computing MEVIS, Bremen, Germany
[6]University of Bremen, Bremen, Germany
*Send Correspond to YG: yi.gao@stonybrookmedicine.edu

## Abstract

Histopathological assessments, including surgical resection and core needle biopsy, are the standard procedures in the diagnosis of the prostate cancer. Current interpretation of the histopathology images includes the determination of the tumor area, Gleason grading, and identification of certain prognosis-critical features. Such a process is not only tedious, but also prune to intra/inter-observe variabilities. Recently, FDA cleared the marketing of the first whole slide imaging system for digital pathology. This opens a new era for the computer aided prostate image analysis and feature extraction based on the digital histopathology images. In this work, we present an analysis pipeline that includes localization of the cancer region, grading, area ratio of different Gleason grades, and cytological/architectural feature extraction. The proposed algorithm combines the human engineered feature extraction as well as those learned by the deep neural network. Moreover, the entire pipeline is implemented to directly operate on the whole slide images produced by the digital scanners and is therefore potentially easy to translate into clinical practices. The algorithm is tested on 368 whole slide images from the TCGA data set and achieves an overall accuracy of 75% in differentiating Gleason 3+4 with 4+3 slides.

## Introduction

Prostate cancer (PCa) is the most common cancer in men in the United States. It is estimated that more than 161,360 new diagnosis is to be made in 2017 with 26,730 death [1]. With wide utilization of serum Prostate Specific Antigen (PSA) testing, a dramatic increase in diagnosis and treatment of PCa has been observed.

Comparing to other leading cancer types, prostate cancer (PCa) shows a favorable long-term prognosis with more than five years of survival after diagnosis[2]. For its diagnosis, the Gleason grading system was developed in the 1960s and still serves as the strongest prognosis predictor[3,4]. It is based on analyzing patterns of glandular and nuclear morphology[4]. The Gleason score consists of two sub-grades: primary grade and secondary grade. The primary grade is assigned to the dominant pattern of the tumor (with greater than 50% in area seen) and secondary grade is assigned to the subordinate pattern. Each of the grades is defined on the scale from 1 to 5, according to the extent of carcinoma infiltration, such as appearance of recognizable glands, with lower grades corresponding to more normal prostate tissue. In the pattern 3, the tissue still has recognizable glands. At high magnification field, part of the cells have left the glands and have trends to infiltrate the surrounding tissue. However, the pattern 4 has fewer recognizable glands, which corresponds to a poorly differentiated carcinoma. Patients diagnosed with Gleason pattern 4 usually need surgical treatments.

Histopathological assessments, including surgical resection and core needle biopsy, are the standard procedures in the diagnosis of the prostate cancer. The study of histology images was regarded as the reference standard to identify disease for diagnosis and treatment, especially cancer grading[5]. However, currently all the diagnoses are made by physicians reviewing the glass slides. Not only is such a process tedious, but also it is prone to intra/inter-observe variabilities.

Recently, FDA cleared the marketing of the first whole slide imaging system for digital pathology. This opens a new era for the inserting the computation components into each diagnosis process. Moreover, the digitization of high-resolution whole slide images (WSIs) makes it possible to implement computer aided diagnosis system to analyze large-scale image data, thus alleviating intra- and inter-observation variations among pathological experts and achieving statistical conclusions across hundreds of slides [6,7]. Specifically in Gleason grading, there is evidence that exact interobserver agreement can be as low as 38%, and 82% for ±1 unit difference [8]. In order to accomplish such a goal, numerous studies have proposed methods for the detection, extraction, and recognition of histopathological patterns [9]. At the nuclear level, algorithms have been proposed to identify and segment nuclei [10–12]. At cellular level, algorithms have been proposed to detect mitosis [13,14] and lymphocytes [15,16]. Moreover, detecting larger structures such as gland has also been actively studied [17–22]. In addition to histology structures, tumor region detection in WSIs also attracts significant attention in the research community [22,23].

Specifically for the PCa, its outcome, in terms of recurrence risk and specific mortality, differs significantly at the different Gleason 7 (intermediate grade) patterns 3+4 and 4+3 after primary therapy [24]. The 10-year prostate cancer specific survival rate for 3+4 (92.1%) was significantly higher than that for 4+3 (76.5%). Considering intermediate Gleason grade 7 accounts for about 40% of all prostate cancers diagnosed [1], the question of accurate choice of the treatment management strategy for those patients becomes of utmost importance. Figure 1 shows two typical WSI of Gleason 7 patterns and the patches extracted from them (Figure 1(c) and Figure 1(d)). The marked area in Figure 1(a) and Figure 1(b) are target areas evaluated during Gleason grading. As can be seen in the corresponding high resolution versions in FIgure 1(c) and 1(d), the gland structures are better differentiated in Grade 3+4 is poorly differentiated in Grade 4+3. In this work, we combine such human defined features with the algorithm identified features into a framework for grading of the intermediate prostate cancer. With accurate and robust classification between Gleason grade 3+4 and 4+3, treatment could be adjusted for better survival.

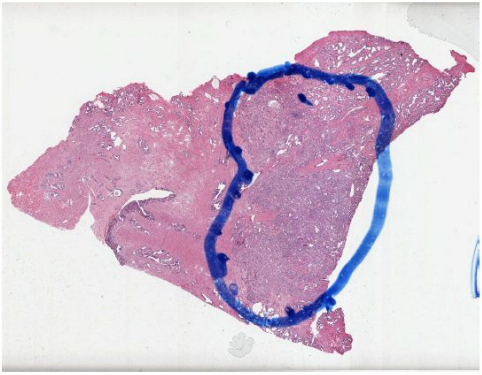 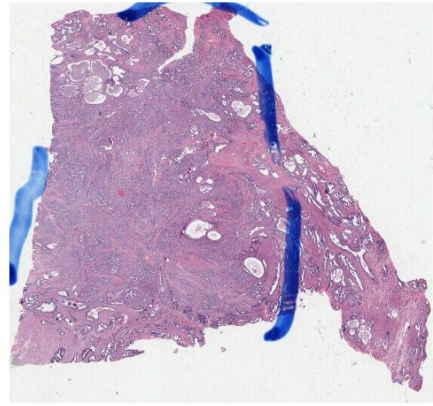

(a) TCGA-EJ-5497, Grade:3+4.  (b) TCGA-EJ-5527, Grade:4+3.

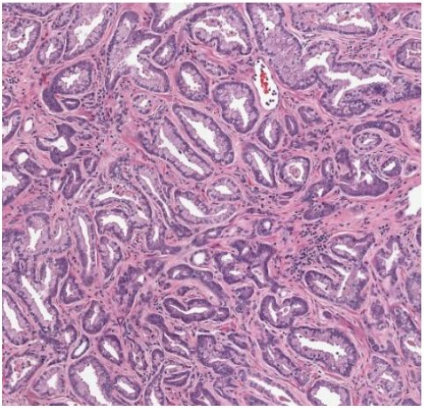 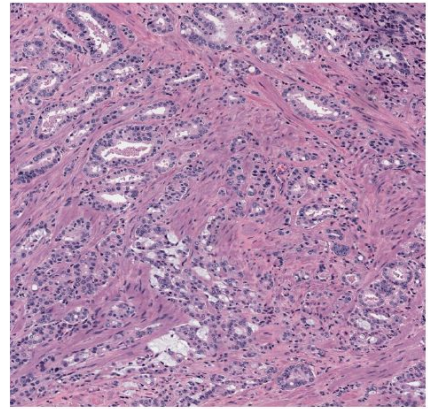

(c) patch of marked area in (a).  (d) patch of marked area in (b).

Figure 1. The intermediate prostate cancer image examples.

For automatic grading of prostate cancer via histopathological image analysis, many state-of-the-art works have been using carefully modeled features based on human observation and assessment, that is, human engineered features.[22,25]. In contrast to human feature engineering, the convolutional neural networks (CNNs) are able to automatically extract features in multiple scales, and simultaneously compute the best separation boundary between the different groups based on the automatically generated features. As fully data driven approaches, CNNs have shown satisfying performance in various computer vision tasks [26]. CNNs have also been applied to medical image analysis and have achieved much success[13,17,18,27]. See the review paper [28] and the references therein.

The Gleason grading system is based on the textural and architectural features of the tumor region. However, a whole slide image consists of inhomogeneous tissue types: malignant gland, benign gland, stroma, etc.). Therefore, in order to achieve higher grading accuracy, one has to first identify the tumor region on the WSI and feed to the classifier. To that end, we first extract the tumor region approximately using the K-means clustering approach. After the tumor regions been defined, the classification is computed based on those regions only. Moreover, the Gleason grading is mainly regarding the nuclear architecture, instead of stromal tissue. The classification is therefore performed in the hematoxylin channel, which captures the nuclear information. As a result, we first perform color decomposition to optimally extracted the hematoxylin channel [29]. Previously, a general color decomposition method based on orthonormal transformation is widely adopted in histopathology analysis research [30]. However, with the variation in the stain condition and scanner, a fixed orthonormal basis could not adaptively adjust the color variation in different slides. In this study, we develop an optimized color decomposition method, for hematoxylin density extraction. After the optimized color decomposition is applied on the tumor region, the hematoxylin channel is then fed to CNN for learning and classification.

One of the major issues for CNN based classification is the preparation for training data. Our approach estimates the overall grade assigned to the slide, without the need to outline individual areas of cancer. We evaluate the accuracy of our grading by comparing the overall grades assigned by the expert with the automatically estimated values.

In addition to the Gleason grades, other features such as prominent nucleoli and cribriform patterns are also found to be related with prognosis. In this framework, such patterns are also captured to generate more complete description of the slide.

# Method

In this section, we first detail our approach for grading the intermediate Gleason patterns. This includes the tumor region extraction, optimized color decomposition, and CNN based classification. After that, the extraction of prognosis significant patterns are described.

## Automatic grading of Gleason 7 patterns

In this section, we detail the classification pipeline. First, using the K-means clustering, we extract the tumor regions so that the subsequent learning and classification are only based on the tumor region, instead of the whole slide. Then, the optimal color decomposition is performed to extract the nuclear information from the tissue. Finally, the CNN is trained on a large set of data for the Gleason 7 (intermediate) grading.

### K-Means Clustering in L*a*b* Color Space

The localized areas (glands) of WSI for further analysis are extracted by K-Means algorithm in L*a*b* color space. The L*a*b* color space is a 3-axis color system with dimension L for lightness and a and b for the color-opponent dimensions. The L*a*b* color space is based on human perception: from human perceptual perspective, the same amount of change in L*a*b* color space produces the same amount of perceptual visual difference. This important property of the La*b* color space provides us the access to using the Euclidean distance in comparing the colors. L (lightness) ranges from 0 to 100. When the value L = 50, it stands for 50% black. The ranges of a and b are both from -128 to +127. In the range, +127a corresponds to pure red, and -128a stands for pure green. Similarly, +127b corresponds to pure yellow, and -128b stands for pure blue. Each color is represented by the combination of the three values.

The values of the two channels (a and b) on each pixel are clustered by K-Means quantization, minimizing the sum of distance functions of each point in the cluster to the K center[31]:

$$s \sum_{i=1}^{k} \sum_{x \in S_i} ||x - \mu_i||^2$$

$(x_1, x_2, ..., x_n)$ observations are partitioned into $k \leq n$, $\mu_i$ is the mean of points in $S_i$.

## Optimized Color Decomposition

### Original Formulation of Color Decomposition

$$O(x,y,c) = -\log\left(\frac{I(x,y,c)+1}{256}\right)$$

where c represents a certain color (red, green or blue). We use OD vectors to describe each stain in the OD-converted RGB color space: $u, v, w \in^3$,
with $\|u\| = \|v\| = \|w\| = 1$, form a matrix $M$ as

$$M = [u, v, w] \in^{3\times 3}$$

for each point $(x, y)$, let $R \in^3$ as $R = O(x, y, :)$. We define the composition of each pure stain in an 8-bit RGB image as a vector $S$. We can get the relationship: $R = M \cdot S$. Then, the decomposition vector $S$ is computed by

$$S = M^{-1} \cdot R$$

### Optimization

A large amount of histopathology image analysis research are conducted on H&E images, which do not contain diaminobenzidine stain. Therefore, the original formulation of color decomposition in Section 4.2.1 is not accurate for hematoxylin and eosin decomposition. To solve this issue, a very direct intuition is to make the third row of matrix $M$ zero. However, in this case, a singular matrix $M$ is non-invertible, thus cause no access to the decomposition vector R (see equation (4)). On the other hand, we also need to control the variation of hematoxylin and eosin channels' variations in $M^{-1}$ to prevent ill-condition. Therefore, in order to obtain a trade-off between the minimization of diaminobenzidine stain channel the variation of matrix $M^{-1}$, we propose a regularized minimization of the sum of diaminobenzidine stain decomposition and the change in matrix $M^{-1}$. In this way, we calculated a new matrix $M^{-1}$ for each whole slide image.

We can directly optimize the matrix $D := M^{-1}$.

Define

$$E(D) := \int\int \|S(3)\|^2 dxdy + \lambda\|D - \bar{D}\|^2$$

$$= \int\int \|DO(x,y,c)\|^2 dxdy + \lambda\|D - \bar{D}\|^2$$

$$= \int\int (d \cdot O(x,y,c))^2 dxdy + \lambda\|D - \bar{D}\|^2$$

where $\bar{D}$ is the original $D$ matrix from experience and
$d$ is the 3rd row of $D$.

Denote $D$ as

$$D = \begin{bmatrix} D_1 & D_2 & D_3 \\ D_4 & D_5 & D_6 \\ D_7 & D_8 & D_9 \end{bmatrix}$$

We use quasi-newton algorithm to minimize $E(D)$ by MATLAB Optimization Toolbox [32].

## Convolutional Neural Network

Convolutional Neural Network (CNN) has achieved tremendous success in computer vision applications. We adopted the CNN architecture from [26], with appropriate modifications for our application. All the layers are constructed as shown in Figure 2. Layers 1 through 6 consist of convolutional (Conv) layers, Rectified Linear Unit (ReLu) activation layers and Max-pooling layers. Layer 7 and 8 are fully connected layers, as in traditional neural networks. The outputs of the Layer 9 are two neurons (two classes), representing Grade 3+3 and Grade 4+4. They are activated by the softmax regression model.

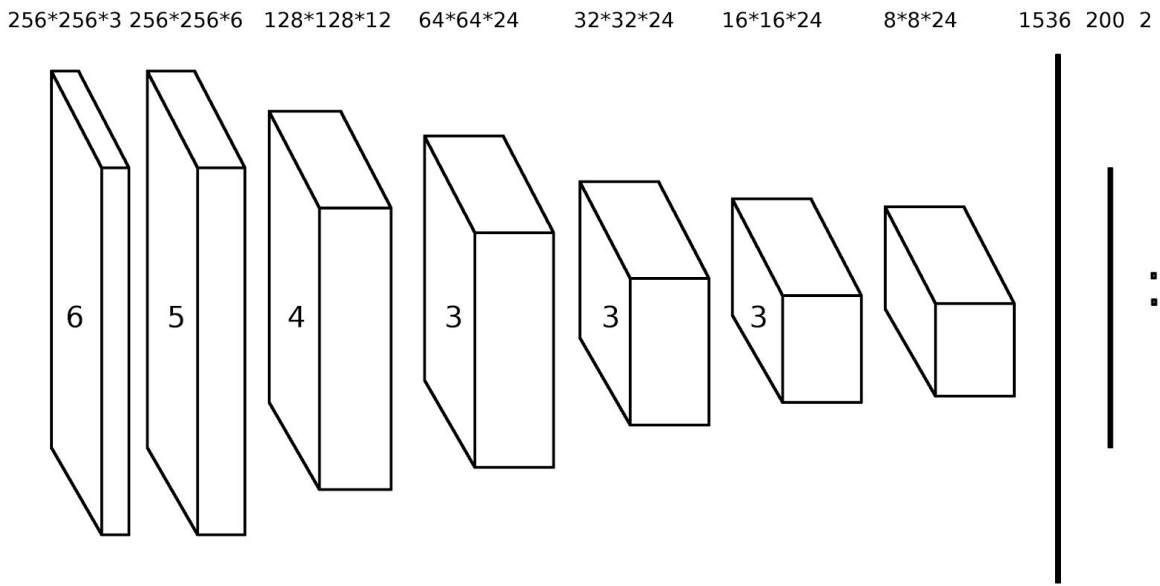

*Figure 2. The convolutional neural network architecture. The number within the layer indicates the kernel size used for the layer. The final two layers fully connected.*

Other network structures including the AlexNet [26], VGG[33], but as will be shown in the result section, the accuracies of these very deep neural network are inferior than the simpler one above.

With input image patches of size 256*256 in the hematoxylin channel obtained from the color decomposition, we collect 100 patches as a batch for training. The drop out ratios in Layer 6 and 7 are both 0.75. Each CNN is trained on GPU for 100,000 iterations (approximately 80 hours on one Nvidia K80 GPU). In the implementation, the TensorFlow framework is used [34]. The ideal network architecture for a task must be tuned via experimentation guided by monitoring the validation set error. Specifically, stochastic gradient descent minimizes the loss function of the mean squared errors during training. The two classes of the training set are roughly numerically balanced, so the issue of overfitting will not be considered in our experiments. Moreover, since the amount of training data is enough for an eight layer CNN, we do not utilize any pre-training techniques.

# Prognosis significant pattern detection

In addition to the traditional Gleason scoring system, several patterns have been identified to be of predictive value for the prognosis. In this work, we design algorithms to detect two patterns.

## Nucleoli prominent

The size of nucleoli is found to be an important indicator of the cancer aggressiveness [35–38] . Figure 3 shows prostate cancer nuclei with prominent nucleoli under 40x objective magnification. In order to aid the diagnosis, the nucleoli prominent nuclei are detected in our pipeline.

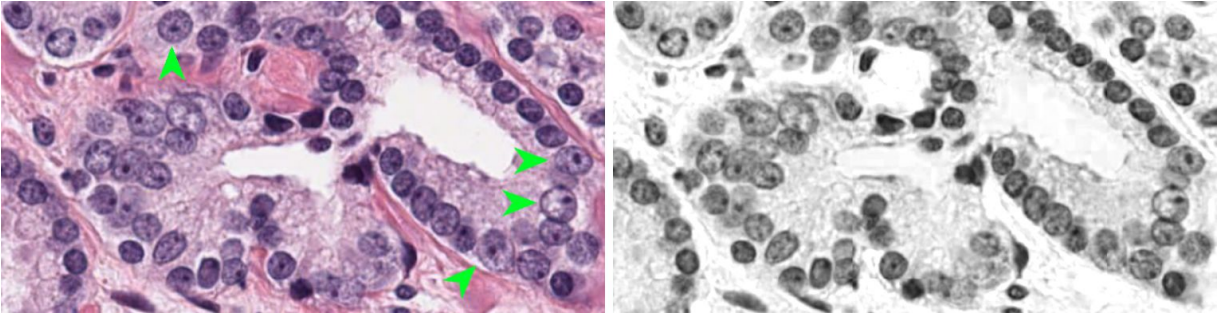

*Figure 3. Prostate cancer at 40x objective magnification. Many nuclei can be seen with prominent nucleoli, some of them indicated by the arrow on the left panel. Right: hematoxylin channel of the image. We see clear nucleoli inside the nucleus*

In order to detect nucleoli, we first follow the algorithm developed in [39]. Then, within each nucleus, histogram is computed in the hematoxylin channel. Nuclei without prominent nucleoli have a single modal Gaussian like distribution. However, nuclei with prominent nucleoli has a significant peak on the dark side of the spectrum. Assuming the intensity follows a Gaussian Mixture Model with two modes, the Expectation-Maximization (EM) algorithm[40] is used to compute the two modes within each nucleus. The mode with lower mean intensity value is considered to be the nucleoli. On the other hand, if the two means are too close (parameter determined empirically), the nucleus is then considered not to have prominent nucleoli.

## Cribriform detection

In the border grades of Gleason 4, there are several tumor growth patterns including fused, ill-defined, cribriform, and glomeruloid glandular structures. It was shown that cribriform strongly predicts distant metastasis and disease-specific death in patients with Gleason score 7 prostate cancer at radical prostatectomy[41]. Patients previously graded as level 3 has now been moved to grade 4 in presence of cribriform. It is therefore important to detect such feature from whole slide image and notify physician. One typical cribriform pattern is shown in Figure 4.

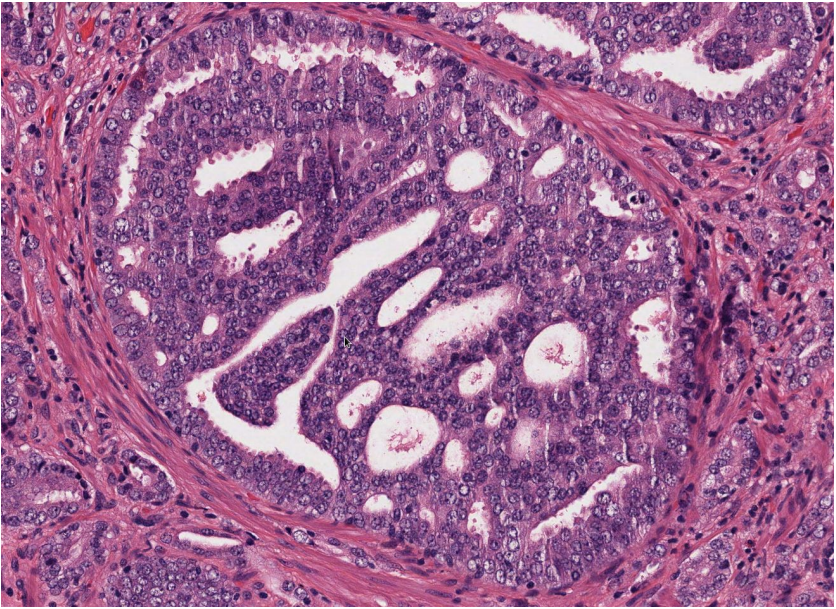

*Figure 4. Example of a cribriform pattern.*

We designed a multi-scale approach to extracted cribriform from whole slide image. In essence, we extract the circular holo feature that appears within a tumor gland. First, the tumor gland region is modeled as regions with high nuclei density and random network characteristics. To that end, the nuclei are extracted using the methods in[39]. Then, a graph is constructed by connecting nuclei with a radius of 30 microns. For each vertex $i$ in the graph, its clustering coefficient $C_i$ is defined as $C_i := \frac{2|e_{jk}:v_j,v_k \in N_i|}{k_i(k_i-1)}$, and is between 0 and 1 [42] . In it, $N_i$ is the neighboring vertices of $i$, $|\cdot|$ denotes the set cardinality, and $k_i = |N_i|$ is the number of neighboring vertexes. For nuclei in the tumor regions, the nuclei are heavily clustered together and their clustering coefficients are high. On the contrary in the stroma region, the value drops to close to 0. In the healthy glandular tissue, however, the clustering coefficient possesses a value in between.

The distribution of the all the cluster coefficients are computed and are separated to three Gaussian modes, using the EM algorithm [40]. The sub-graph with highest clustering coefficients are regarded as tumor gland regions.

Cribriform glands are characterized by the circular hollow region in the gland. To that end, all the non-background hollow regions are extracted with the simple criteria of having all the three color channels (red, green, and blue) larger than 200. These regions then intersects with the tumor subgraph region detected above. For each of them, the roundness of the region is computed as $R = A/a$ following[43], where $a$ denotes the area of a region and $A$ is the area of the disk with the same circumference. If the region is a perfect circle, $R = 1$ and the more irregular it becomes, the closer it approaches 0. Finally, regions with roundness above 0.7 are kept.

## Evaluation

The training of CNN typically requires large amount of labeled datasets. Digitized machine readable annotations are, unfortunately, not available in typical datasets obtained in the clinical setting due to the difficulty in recruiting well trained medical personnel to perform tedious labeling work. In our study, we approach evaluation of the proposed automatic grading pipeline using the approach detailed below. Since our main interest lies in the differentiation between Gleason grade 3+4 and 4+3, we use as our training data all the digitized slides with the expert-assigned overall Gleason grade 3+3 for one class and Gleason grade 4+4 for the other. We assume that all the tumor region extracted by the algorithm above from Gleason 3+3 (4+4, resp.) are "pure" Gleason pattern 3 (4, resp.). By taking our approach, we avoid the time consuming contouring task. This enables the opportunity of using much large dataset for training the CNN for optimal classification performance. For the evaluation of the automated classification, we use the processing pipeline to assign the overall Gleason grad to a given slide, and compare the result with the expert-assigned grade.

On the higher grade end, in addition to Gleason 4+4, we also include the slides with grades higher than 4+4 for the training. As a result, the trained classifier is indeed differentiating grade 3 against grades 4 and above. Based on such reasoning, the slides with Gleason grade 2+4, 3+5, 5+3 are discarded because one can be certain about whether a patch extracted from them is of grade 3, or grade 4 and above.

Given a WSI, the patches on its tissue region is extracted randomly and classified by the trained CNN. If more patches are graded as level 3 than level 4, the overall WSI is graded as Gleason grade 3+4, and 4+3 otherwise. Indeed, based on the design of the classifier, the classifier is only able to detect class 4 and above. Therefore the resulting class assignment can be interpreted as 3+4* or 4*+3, where 4* indicates "4 and above".

# Results

## Tumor region extraction

Figure 4 shows the K-means based tumor region extraction.

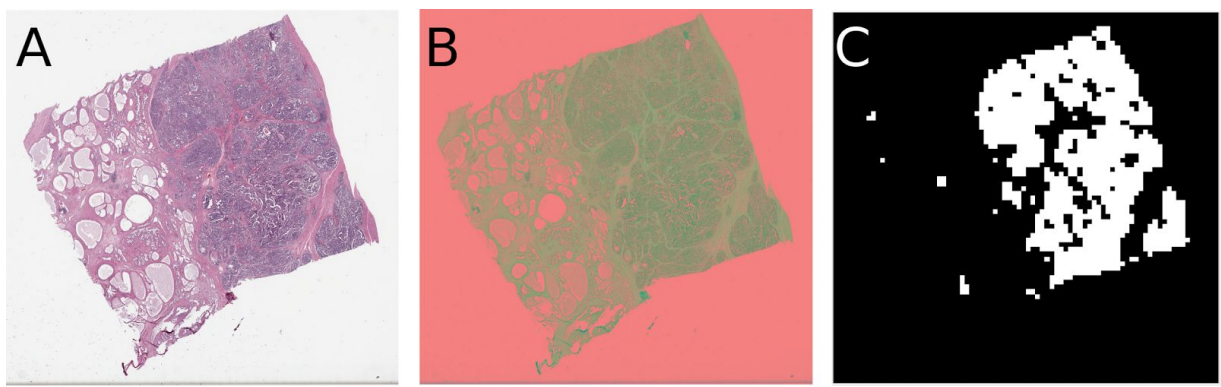

*Figure 5. A. Original WSI. B. WSI in L\*a\*b color space. C. Tumor region mask*

The lowest resolution level of each WSI is converted into L*a*b* color space, as shown in Figure 5A and 5B. Using K-means algorithm we extract 3 groups from each WSI. In some cases, there are pen markers on the slides (such as the blue markers in Figure 1A and 1B). While those markers were intended to circle out certain region of interest, they are not used for consistency across all data set. Instead, 4 groups are extracted from those slides. We select the cluster with second maximum mean value in blue channel as the tumor region, as shown in Figure 5C. The masked images are generated to localize rumor areas for further color decomposition and classification.

## Optimal color decomposition for H&E slides

Figure 5 shows the optimized color decomposition results. In both rows, the pre-defined hematoxylin color components causes the resulting hematoxylin channel to be over-saturated. Consequently, the cytoplasm region is easy to be confused with nuclear region. This is more prominent in the higher grade, such as the Grade 4 pattern in Figure 6(e). In contrast, the optimized color decomposition yields a clearer extraction of the hematoxylin component, shown in Figure 6(f). This makes the subsequent nuclei extraction more accurate.

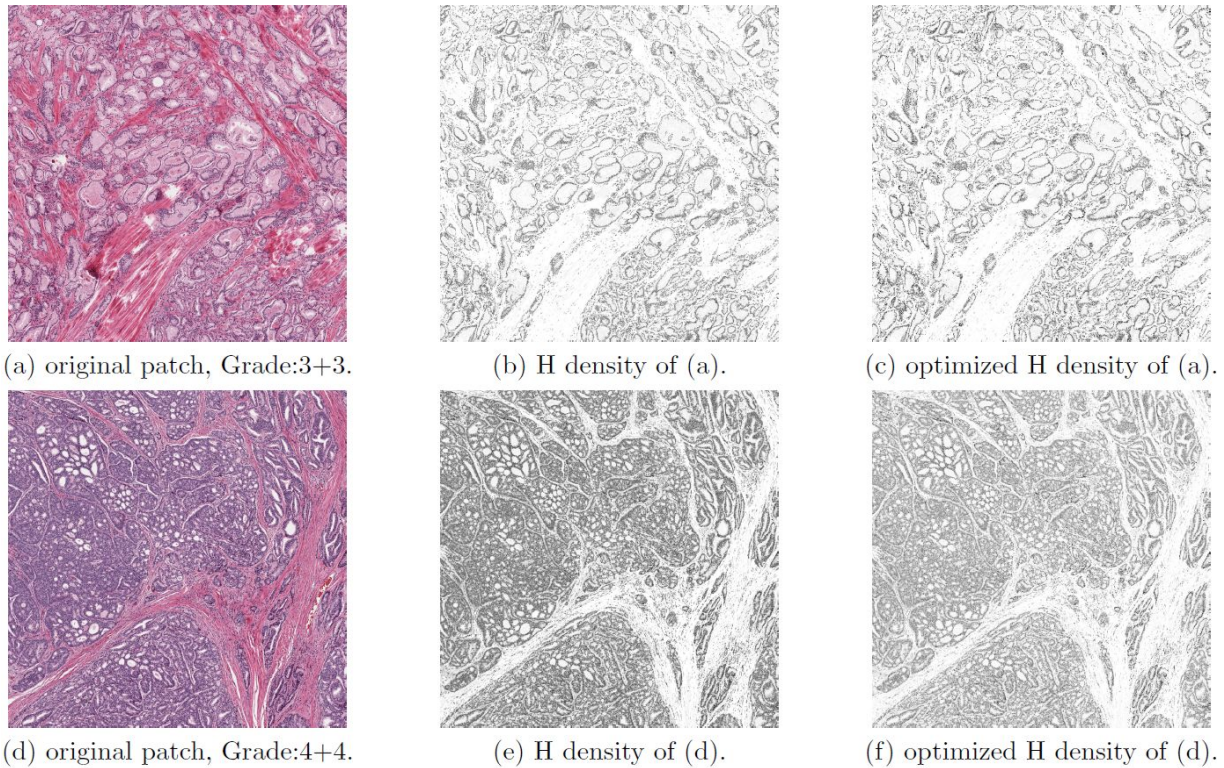

Figure 6. The original and optimized color decomposition results.

## Automatic grading and detection of Cribriform

Our evaluation utilizes the publicly available Prostate Adenocarcinoma dataset [44,45] from The Cancer Genome Atlas (TCGA) [46], which includes 368 digitized prostate pathology slides.

A total of 495 subjects available on TCGA were utilized for the evaluation of automatic Gleason grading. Among them 380 slides are available. The distribution of the overall Gleason grades assigned by the experts manually used as the reference are shown in Table 1.

> Table 1. Number of cases for each of the individual Gleason grades from the Prostate Adenocarcinoma TCGA collection used in this study. Cells highlighted in red correspond to the cases that were not used in the evaluation, since they intermix low and high grade Gleason

| patterns. Cases highlighted in blue are used for training CNN classifier, which are applied on the cases highlighted in yellow. | | | | | | | | | |
|---|---|---|---|---|---|---|---|---|---|
| Glease grades | 2+4 | 3+3 | 3+4 | 4+3 | 4+4 | 4+5 | 5+4 | 5+3 | 3+5 | 5+5 |
| # cases | 1 | 38 | 114 | 76 | 47 | 74 | 16 | 6 | 5 | 3 |

The WSI level overall classification accuracy was 75% based on the 368 slides suitable for the analysis, as described in Table 1. Figure 7A shows the contour of applying the classifier on one WSI. In this case, we display the boundary of patches classified as level 4 in blue and level 3 in green.

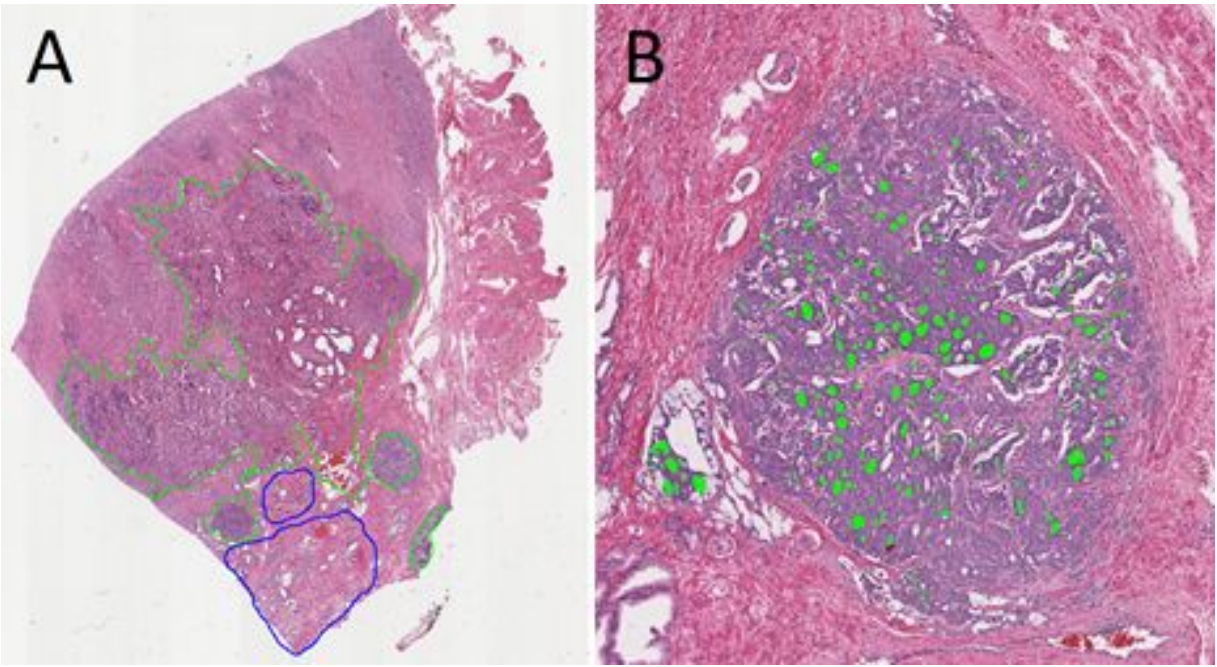

Figure 7. (A) Algorithm delineated Gleason grade 4 (blue) and 3 (green) region contours. Sub-figure (B) zooms into one of the level 4 region highlighting the automatically detected cribriform pattern.

## Limitations

Our work has limitations. Specifically, we did not have annotations of the tumor regions for the slides being used in the evaluation. As such, we were not able to evaluate agreement of the automatically identified tumor areas (e.g., those shown in Fig. 4). Future work for evaluation of the proposed methodology will need to concentrate on contrasting the automatically identified areas with the expert annotations.

The deep learning classification framework assumes that the image patch belongs to either of the two Gleason grades considered. Therefore, if a patch has benign or aggressive cancer present, we cannot properly address this situation. The see the main application of the classification system in precisely quantifying the individual areas of Gleason 3 and 4 in intermediate grade cancer areas. We envision that in the future such intermixed areas can be manually annotated, leaving precise annotations for automated processing.

# Conclusions

In this paper, we have introduced combine feature extraction and deep learning based pipeline for prostate whole slide histopathology image analysis.. The workflow consists of gland region segmentation by K-Means clustering, optimized color decomposition of hematoxylin channel, convolutional neural network based classification, and prognosis related characteristics extraction. Moreover, we used a label-free approach for grading of the intermediate prostate cancer without handcrafted ground truth, which enables training in much larger scale possible.

# References

1. Siegel RL, Miller KD, Jemal A. Cancer Statistics, 2017. CA Cancer J Clin. 2017;67: 7–30.

2. Daneshmand S, Quek ML, Stein JP, Lieskovsky G, Cai J, Pinski J, et al. Prognosis of patients with lymph node positive prostate cancer following radical prostatectomy: long-term results. J Urol. 2004;172: 2252–2255.

3. Mellinger GT, Gleason D, Bailar J 3rd. The histology and prognosis of prostatic cancer. J Urol. mysciencework.com; 1967;97: 331–337.

4. Epstein JI. An update of the Gleason grading system. J Urol. Elsevier; 2010;183: 433–440.

5. Elston CW, Ellis IO, Others. Pathological prognostic factors in breast cancer. I. The value of histological grade in breast cancer: experience from a large study with long-term follow-up. Histopathology. Wiley Online Library; 1991;19: 403–410.

6. Ghaznavi F, Evans A, Madabhushi A, Feldman M. Digital imaging in pathology: whole-slide imaging and beyond. Annu Rev Pathol: Mech Dis. Annual Reviews; 2013;8: 331–359.

7. Commissioner O of T. FDA allows marketing of first whole slide imaging system for digital pathology. Office of the Commissioner; Available: https://www.fda.gov/NewsEvents/Newsroom/PressAnnouncements/ucm552742.htm. Accessed 14 Apr 2017.

8. Harada M, Mostofi FK, Corle DK, Byar DP, Trump BF. Preliminary studies of histologic prognosis in cancer of the prostate. Cancer Treat Rep. europepmc.org; 1977;61: 223–225.

9. Gurcan M n., Boucheron L e., Can A, Madabhushi A, Rajpoot N m., Yener B. histopathological image analysis: a review. IEEE Rev Biomed Eng. ieee; 2009;2: 147–171.

10. Xing F, Yang L. Robust Nucleus/Cell Detection and Segmentation in Digital Pathology and Microscopy Images: A Comprehensive Review. IEEE Rev Biomed Eng. ieeexplore.ieee.org; 2016;9: 234–263.

11. Irshad H, Veillard A, Roux L, Racoceanu D. Methods for nuclei detection, segmentation, and classification in digital histopathology: a review—current status and future potential. IEEE Rev Biomed Eng. IEEE; 2014;7: 97–114.

12. Li G, Raza SEA, Rajpoot NM. Multi-resolution cell orientation congruence descriptors for epithelium segmentation in endometrial histology images. Med Image Anal. Elsevier; 2017;37: 91–100.

13. Wang H, Cruz-Roa A, Basavanhally A, Gilmore H, Shih N, Feldman M, et al. Mitosis detection in breast cancer pathology images by combining handcrafted and convolutional neural network features. J Med

Imaging (Bellingham). 2014;1: 034003.

14. Cireşan DC, Giusti A, Gambardella LM, Schmidhuber J. Mitosis detection in breast cancer histology images with deep neural networks. J Med Imaging Radiat Oncol. Springer; 2013;1: 411–418.

15. Heindl A, Nawaz S, Yuan Y. Mapping spatial heterogeneity in the tumor microenvironment: a new era for digital pathology. Lab Invest. nature.com; 2015;95: 377–384.

16. Zhao T, Hou L, Nguyen V, Gao Y, Samaras D, Kurc T, et al. Using Machine Methods to Score Tumor-Infiltrating Lymphocytes in Lung Cancer. LABORATORY INVESTIGATION. NATURE PUBLISHING GROUP 75 VARICK ST, 9TH FLR, NEW YORK, NY 10013-1917 USA; 2017. p. 403A–403A.

17. Cruz-Roa A, Basavanhally A, González F, Gilmore H, Feldman M, Ganesan S, et al. Automatic detection of invasive ductal carcinoma in whole slide images with convolutional neural networks. SPIE Medical Imaging. International Society for Optics and Photonics; 2014. pp. 904103–904103–15.

18. Cruz-Roa AA, Arevalo Ovalle JE, Madabhushi A, González Osorio FA. A deep learning architecture for image representation, visual interpretability and automated basal-cell carcinoma cancer detection. Med Image Comput Comput Assist Interv. 2013;16: 403–410.

19. Lee G, Sparks R, Ali S, Shih NNC, Feldman MD, Spangler E, et al. Co-Occurring Gland Angularity in Localized Subgraphs: Predicting Biochemical Recurrence in Intermediate-Risk Prostate Cancer Patients. IEEE Trans Med Imaging. IEEE; 2014;32: 1804–1818.

20. Sirinukunwattana K, Pluim JPW, Chen H, Qi X, Heng P-A, Guo YB, et al. Gland segmentation in colon histology images: The glas challenge contest. Med Image Anal. Elsevier; 2017/1;35: 489–502.

21. Chen H, Qi X, Yu L, Heng P-A. Dcan: Deep contour-aware networks for accurate gland segmentation. Proceedings of the IEEE conference on Computer Vision and Pattern Recognition. cv-foundation.org; 2016. pp. 2487–2496.

22. Nguyen K, Sabata B, Jain AK. Prostate cancer grading: Gland segmentation and structural features. Pattern Recognit Lett. 2012;33: 951–961.

23. He L, Long LR, Antani S, Thoma GR. Histology image analysis for carcinoma detection and grading. Comput Methods Programs Biomed. 2012;107: 538–556.

24. Wright JL, Salinas CA, Lin DW, Kolb S, Koopmeiners J, Feng Z, et al. Prostate cancer specific mortality and Gleason 7 disease differences in prostate cancer outcomes between cases with Gleason 4+ 3 and Gleason 3+ 4 tumors in a population based cohort. J Urol. Elsevier; 2009;182: 2702–2707.

25. Ginsburg SB, Lee G, Ali S, Madabhushi A. Feature Importance in Nonlinear Embeddings (FINE): Applications in Digital Pathology. IEEE Trans Med Imaging. 2016;35: 76–88.

26. Krizhevsky A, Sutskever I, Hinton GE. ImageNet Classification with Deep Convolutional Neural Networks. In: Pereira F, Burges CJC, Bottou L, Weinberger KQ, editors. Advances in Neural Information Processing Systems 25. Curran Associates, Inc.; 2012. pp. 1097–1105.

27. Xu Y, Jia Z, Ai Y, Zhang F, Lai M, Chang EIC. Deep convolutional activation features for large scale Brain Tumor histopathology image classification and segmentation. 2015 IEEE International Conference on Acoustics, Speech and Signal Processing (ICASSP). 2015. pp. 947–951.

28. Shen D, Wu G, Suk H-I. Deep Learning in Medical Image Analysis. Annu Rev Biomed Eng. annualreviews.org; 2017; doi:10.1146/annurev-bioeng-071516-044442

29. Avwioro G. Histochemical uses of haematoxylin—a review. JPCS. 2011;1: 24–34.

30. Ruifrok AC, Johnston DA. Quantification of histochemical staining by color deconvolution. Anal Quant Cytol Histol. 2001;23: 291–299.

31. MacQueen J, Others. Some methods for classification and analysis of multivariate observations. Proceedings of the fifth Berkeley symposium on mathematical statistics and probability. Oakland, CA, USA.; 1967. pp. 281–297.

32. Shanno DF. Conditioning of quasi-Newton methods for function minimization. Math Comput. 1970;24: 647–656.

33. Simonyan K, Zisserman A. Very Deep Convolutional Networks for Large-Scale Image Recognition [Internet]. arXiv [cs.CV]. 2014. Available: http://arxiv.org/abs/1409.1556

34. Abadi M, Agarwal A, Barham P, Brevdo E, Chen Z, Citro C, et al. TensorFlow: Large-Scale Machine Learning on Heterogeneous Distributed Systems [Internet]. arXiv [cs.DC]. 2016. Available: http://arxiv.org/abs/1603.04467

35. Busch H, Byvoet P, Smetana K. The nucleolus of the cancer cell: a review. Cancer Res. AACR; 1963; Available: http://cancerres.aacrjournals.org/content/23/3/313.short

36. Nguyen K, Jain AK, Sabata B. Prostate cancer detection: Fusion of cytological and textural features. J Pathol Inform. 2011;2: S3.

37. Yap CK, Kalaw EM, Singh M, Chong KT, Giron DM, Huang C-H, et al. Automated image based prominent nucleoli detection. J Pathol Inform. ncbi.nlm.nih.gov; 2015;6: 39.

38. Derenzini M, Trerè D, Pession A, Govoni M, Sirri V, Chieco P. Nucleolar size indicates the rapidity of cell proliferation in cancer tissues. J Pathol. 2000;191: 181–186.

39. Zhou N, Yu X, Zhao T, Wen S, Wang F, Zhu W, et al. Evaluation of nucleus segmentation in digital pathology images through large scale image synthesis. SPIE Medical Imaging. International Society for Optics and Photonics; 2017. p. 101400K–101400K.

40. Dempster AP, Laird NM, Rubin DB. Maximum Likelihood from Incomplete Data via the EM Algorithm. J R Stat Soc Series B Stat Methodol. [Royal Statistical Society, Wiley]; 1977;39: 1–38.

41. Kweldam CF, Wildhagen MF, Steyerberg EW, Bangma CH, van der Kwast TH, van Leenders GJLH. Cribriform growth is highly predictive for postoperative metastasis and disease-specific death in Gleason score 7 prostate cancer. Mod Pathol. nature.com; 2015;28: 457–464.

42. Watts DJ, Strogatz SH. Collective dynamics of "small-world"networks. Nature. Nature Publishing Group; 1998;393: 440–442.

43. Lehmann G. Label object representation and manipulation with ITK. Insight J. insight-journal.org; 2007;8. Available: http://insight-journal.org/midas/item/view/1343

44. Prostate Adenocarcinoma. In: The Cancer Genome Atlas - National Cancer Institute [Internet]. 2015 [cited 5 May 2017]. Available: https://cancergenome.nih.gov/cancersselected/prostatecancer

45. Cancer Genome Atlas Research Network. The Molecular Taxonomy of Primary Prostate Cancer. Cell. Elsevier; 2015;163: 1011–1025.

46. Weinstein JN, Collisson EA, Mills GB, Shaw KRM, Ozenberger BA, Ellrott K, et al. The cancer genome atlas pan-cancer analysis project. Nat Genet. Nature Publishing Group; 2013;45: 1113–1120.